\documentclass{article}
\usepackage{spconf,amsmath,graphicx}
\usepackage{epsfig}
\usepackage{amssymb}
\usepackage{subfigure}
\usepackage{cite}
\usepackage{multirow}
\usepackage{epstopdf}
\usepackage{graphicx}
\usepackage{amsmath}
\usepackage{color, soul}
\usepackage[style=base]{caption}
\usepackage{lineno}
\usepackage{comment}
\usepackage{array}  
\usepackage{siunitx}
\usepackage{mathtools}
\usepackage{calrsfs}
\usepackage{eucal}
\usepackage{bm}
\newcolumntype{d}{>{\displaystyle}c}

\usepackage{url}
\usepackage{hyperref}
\hypersetup{
	colorlinks=true,       
	linkcolor=red,          
	citecolor=green,        
	filecolor=magenta,      
	urlcolor=cyan           
}
\usepackage{enumitem}
\usepackage{pgfplots}

\usepackage{array}
\newcolumntype{L}[1]{>{\raggedright\let\newline\\\arraybackslash\hspace{0pt}}m{#1}}
\newcolumntype{C}[1]{>{\centering\let\newline\\\arraybackslash\hspace{0pt}}m{#1}}
\newcolumntype{R}[1]{>{\raggedleft\let\newline\\\arraybackslash\hspace{0pt}}m{#1}}

\title{A deep learning approach for analyzing the composition of chemometric data}
\name{
	\begin{tabular}{ccc}
		Muhammad Bilal$^1$ & Mohib Ullah$^2$  
	\end{tabular}
}
\address{
	\begin{tabular}{c}
		$^1$University of Trento, Italy. \\
		$^2$Norwegian University of Science and Technology, Norway. 
	\end{tabular}
}

\begin{document}
%
\maketitle

\begin{abstract}
	We propose novel deep learning based chemometric data analysis technique. We trained L2 regularized sparse autoencoder end-to-end for reducing the size of the feature vector to handle the classic problem of the curse of dimensionality in chemometric data analysis. We introduce a novel technique of automatic selection of nodes inside the hidden layer of an autoencoder through Pareto optimization. Moreover, Gaussian process regressor is applied on the reduced size feature vector for the regression. We evaluated our technique on orange juice and wine dataset and results are compared against 3 state-of-the-art methods. Quantitative results are shown on Normalized Mean Square Error (NMSE) and the results show considerable improvement in the state-of-the-art. 
\end{abstract}

\begin{figure*}
	\centering
	\includegraphics[scale=0.36]{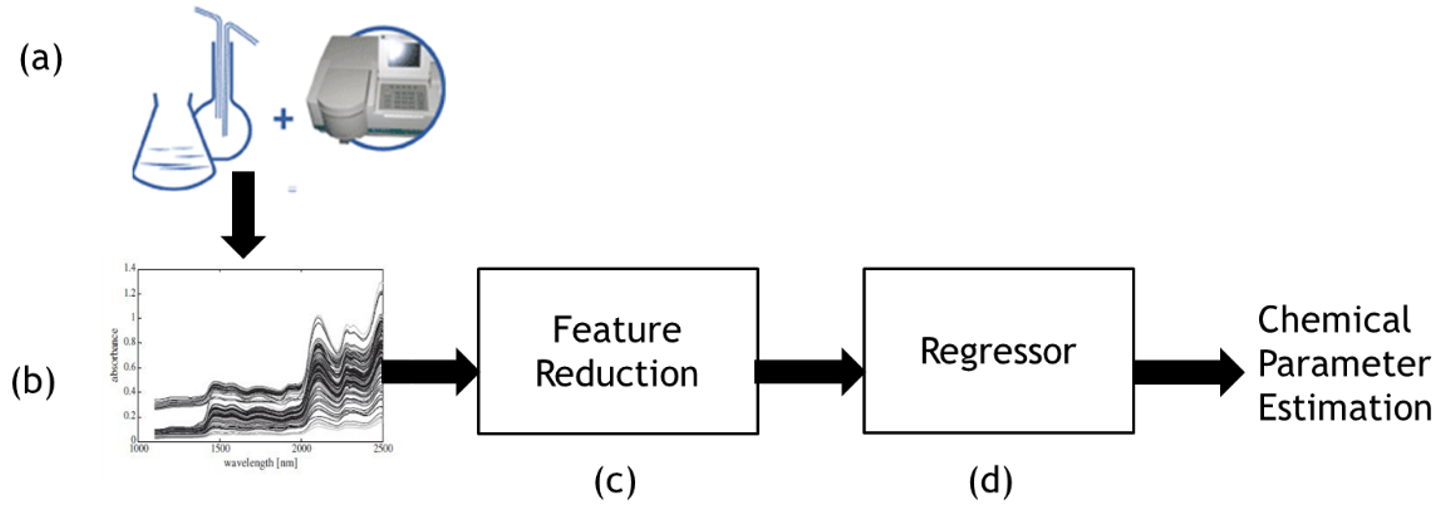}
	\caption{(a) Data collection through spectrophotometry.  
		(b) Spectral signature.
		(c) Feature reduction through autoencoder neural network. 
		(d) Gaussian process regressor for estimating chemical parameter of interest. }
	\label{PL}
\end{figure*}

\section{Introduction} 
\label{sec:intro}

Chemometrics or multivariate data analysis is the science which applies statistical and mathematical methods to process the data obtained through spectroscopic techniques, in order to derive information of interest. The need for chemometric analysis comes from the development of analytical instruments and techniques that are capable of producing large amount of complex data. Data collection through spectroscopic technique is based on interaction of light energy of variable wavelength with samples under test \cite{wold1995chemometrics}. The ability of a sample to absorb or transmit light energy is recorded in terms of values throughout a selected bandwidth of electromagnetic spectrum. Whether it be food, pharmaceutical or textile industry, concentrations of chemical components of interest in samples are estimated through chemometric analysis. Spectroscopic data often comprises of more spectral features or variables than the samples or observations. This not only creates curse of dimensionality issue but also the consecutive variables in a spectrum are highly correlated in nature, that is, some spectral variables can be represented as linear combination of other independent variables. The work of Bertrand et al. \cite{bertrand2002spectroscopie} and Sutter et al. \cite{sutter1993comparison} have shown that existence of such high collinearity between the spectral variables can result in inaccurate predictions. It is problematic to apply directly statistical methods due to high collinearity, like multiple linear regression (MLR) in \cite{belsley2005regression, eklov1999selection, geladi2002some, naes2001understanding}.

In order to deal with data dimensionality and data redundancy, different works have been proposed in literature \cite{verleysen2003learning}. Most of the previous work suggests reducing the number of variables or features to cope with curse of dimensionality problem, thus allowing to get more accurate results through regression techniques. The feature reduction can be generally achieved in two different methods. The first one contains selecting the most relevant features based on a chosen criterion from the original set of features \cite{cornell1987classical, benoudjit2004chemometric, benoudjit2004spectrophotometric, rossi2006mutual}, while in second case the original features are transformed from one space to another in such a way to keep the reconstruction error as minimum as possible. This transformation of features set from one space to another can either be linear or non-linear depending upon the scope of application. The examples of later method are Principle Component Regression \cite{wold1987principal} (PCR) and Partial Least Square Regression (PLSR) \cite{geladi2002some}. PCR is composed of simple linear regression model based on few principle components of the original spectral data. While PLSR focuses on calculating the linear projections that shows maximum correlation with the output or target variable, thus estimating a linear regression model determined by the projected coordinates. Benoudjit et al. \cite{benoudjit2004chemometric} proposed linear and nonlinear regression methodologies which are based upon an incremental routine for feature selection and using a validation set. In \cite{benoudjit2004spectrophotometric, rossi2006mutual} different techniques have been introduced to improve the results of previous method by choosing the best feature set for initializing the routine and finding a feature selection strategy that depends entirely on the shared information between spectral data and target variable. An interesting approach to the chemometrics problems has been discussed in \cite{benoudjit2009multiple}, where instead of traditional feature reduction approach, the whole information in spectral data space is exploited by using Multiple Regression System (MRS). Similarly, Douak et al.\cite{douak2011two} come up with two stage regression approach that is based on residual-based correction (RBC) concept. Their basic idea is to correct any adopted regressor, called functional estimator, by analyzing and modeling its residual errors directly in feature space. The results obtained from MRS technique outperform the ones obtained through traditional feature reduction techniques such as PCR and PLSR \cite{geladi2002some, benoudjit2009multiple}.

The importance of salient features have been highlighted in other computer vision applications like scene understanding \cite{ullah2012crowd, ullah2013structured, ullah2016crowd}, crowd analysis \cite{ullah2017density, rota2013particles, ullah2015crowd}, illuminant estimation \cite{ahmad2017illumination}, segmentation \cite{ullah2018hybrid, ullah2014dominant}, and anomaly detection \cite{ullah2018anomalous}, to name a few. Similarly, In the last few years, deep learning has shown outstanding results on a image classification \cite{szegedy2017inception}, segmentation \cite{ullah2018pednet}, and tracking \cite{ullah2018deep, ullah2018directed} etc. Inspired by the success of deep learning in such applications, we proposed an automatic deep learning based feature extraction technique. In a nutshell, deep models learns hierarchical features \cite{ullah2017hierarchical, ullah2019internal} and have the capability to learn the structure of the underlying data. We designed an autoencoder neural network \cite{baldi2012autoencoders} for removing redundant and irrelevant features from the spectral data. It is an automatic feature extraction technique that is capable of linear and nonlinear feature extraction based upon the selection of parameters of the architecture. Pareto optimization technique is applied in order to choose the best architecture (in terms of model complexity) of autoencoder for the feature extraction. After extracting meaningful features from the original feature set, we exploited Gaussian process regression to solved our regression problem.

The rest of the paper is organized in the following order. In the section \ref{PA}, we explain our proposed method. In section \ref{GM}, we present autoencoder training and parameter optimization. Section \ref{GPR} illustrate our baseline regressor. In section \ref{EX}, quantitative results are presented and section \ref{CN} concludes the paper. 

\section{Proposed Approach}
\label{PA}
We focus on estimating sugar and alcohol concentration in orange juice and wine datasets. The whole pipeline can be seen in 3 discrete steps. In the first step, date is acquired from the liquids through near infrared (orange juice data set) and mid infrared (wine dataset) reflectance spectroscopy technique Fig. \ref{PL}(a,b). In the 2nd step, the acquired data is processed through auto-encoder neural network for feature reduction Fig. \ref{PL} (c) and then in the last step, Gaussian process regressor estimate the concentrations of mentioned components Fig. \ref{PL} (d). The aim of autoencoder is to retrieve set of features which are the best representation of original data without redundancy. In the next section \ref{GM}, the feature reduction step is explained. 
\section{Autoencoder}
\label{GM}
An autoencoder is a neural network that is trained to reconstruct the input data into the output with minimum amount of reconstruction error. They are designed in a way not to copy the input but to learn important and unique feature of the input data. Autoencoders are mainly used for pre-train deep networks dimensionality reduction, feature learning and generative modeling of data. It is composed of two main parts, input layer and output layer together with hidden layer connecting the two layers. The input layer has the same number of nodes as the output layer. To build an autoencoder we need encoding function at the input, decoding function at the output and loss function to calculate the amount of information loss between encoded representation and decoded representation of the input and output data respectively.
\subsection{Encoder}
It maps an input vector $x \in \mathbb{R}^n$, into encoded representation $h(x) \in \mathbb{R}^m$. The typical form is affine mapping followed by nonlinearity (eq. \ref{ac}). The parameter is set $\Theta = (w,b)$ where $w$ is weight matrix of size $mxn$ and $b \in \mathbb{R}^m$ is bias vector, $f$ is activation function.
\begin{equation}
\label{ac}
h_{\Theta} = f(wx+b)
\end{equation}

\subsection{Decoder}
It maps the resulting encoded representation $h(x)$ back into an estimate of reconstructed n-dimensional vector $r \in \mathbb{R}^n$, where
\begin{equation}
r_{\phi} = g(f(x))
\end{equation}

\begin{equation}
r_{\phi} = g(w'h+b')
\end{equation}

The parameter $\phi = \big\{w',b'\big\}$, where $w'$ is weight matrix of size is $nxm$, $b' \in \mathcal{R}^n$ is bias vector and $g$ is activation function of the decoder. The autoencoder tries to learn a function $r_{\phi} \simeq x$ , thus each training data $x(i)$ is mapped to corresponding reconstructed data $r(i)$.

\subsection{Loss function}
Imposing sparsity constraint on the hidden nodes of an autoencoder enables the model to learn unique statistical features of the data even when the number of hidden units are larger compared to the feature space of the input data. A neuron is considered active if its output value is close to maximum value of the activation function used (close to 1 for sigmoid activation function) and inactive if its output is close to minimum value (0 in case of sigmoid). The average activation of $i^{th}$ hidden unit $\hat{\rho_i}$ (eq. \ref{par1}), where $n$ is total number of inputs, $x_j$ is the $j^{th}$ training example and $h_i$ activation of $j^{th}$ hidden unit is given as:

\begin{equation}
\label{par1}
\hat{\rho_i} = \frac{1}{n} \sum_{j=1}^{n} h_i(x_j)
\end{equation}

Choosing sparsity parameter $\rho$ small ($\rho$=0.01) and imposing $\hat{\rho_i}=\rho$  constraint, Kullback-Leibler divergence term is applied to penalize $\hat{\rho_i}$. Penalty value that diverges from $\rho$ will give reasonable result. 

\begin{equation}
KL(\rho||\hat{\rho_i}) = \rho \log(\frac{\rho}{\hat{\rho_i}}) + (1-\rho) \log (\frac{1-\rho}{1-\hat{\rho_i}})
\end{equation}

where $KL(\rho||\hat{\rho_i})$ is the Kullback-Leibler divergence between Bernoulli random variable with $\rho$ mean and Bernoulli random variable with mean $\hat{\rho_i}$. If $\rho = \hat{\rho_i}$, $KL(\rho||\hat{\rho_i}) = 0$, else increases monotonically as $\hat{\rho_i}$ diverges from $\rho$. Minimizing the $KL$ divergence penalty term leads to $\hat{\rho_i}$ to be close to $\rho$. The overall loss function will be as the following: \newline

${\scriptstyle L(\Theta, \Omega) = \frac{1}{N} \sum_{n=1}^{N} \sum_{k=1}^{K} (x_{kn} - r_{kn})^2 + \lambda * \Omega_{weights} + \beta * \Omega_{sparsity}}$\newline 

where $\Omega_{sparsity}$ is sparsity regularizer and calculated as eq. \ref{par2}:

\begin{equation}
\label{par2}
\Omega_{sparsity} = \sum_{i=1}^{m} KL(\rho||\hat{\rho_i})
\end{equation}

Similarly, $\Omega_{weights}$ is $L2$ regularization term. It's task is to avoid overfitting by penalizing the rate in which the model reacts to changes in the training example distribution and forcing the model to learn most significant features. It is calculated as eq. \ref{par3}
\begin{equation}
\label{par3}
\Omega_{weights} = \frac{1}{2} \sum_{l}^{L} \sum_{j}^{N} \sum_{i}^{K} (w_{ij}^{l}))^2
\end{equation}
where $L$ is number of hidden layers, $N$ is the number of training examples and $K$ is the number of features. In the lost function, $\lambda$ is coefficient for the $L2$ regularization term $\Omega_{weights}$ and $\beta$ is the coefficient for sparsity regularization $\Omega_{sparsity}$ term.

\subsection{Training}
\label{DFEC}
Once the model is setup, our goal is to minimize the cost function $L(\Theta, \Omega)$ as a function of weights $w$ and bias $b$. To train our autoencoder neural network, we initialized each parameter $w_{i,j}^{(l)}$,and $b_{i}^{(l)}$ each to a small random value near zero $($ $\mathcal{N}(0,\varepsilon^2)$ distribution for a small $\varepsilon$)$)$, and then apply stochastic conjugate gradient decent (SCG) algorithms to learn the network parameters of autoencoder. Random initialization is necessary, if all the parameters start off at identical values, then all the hidden layer units will end up learning the same function of the input.

\subsubsection{Pareto Based Multi Objective Learning(PMOL)} 
The choice on model complexity and mean squared error (MSE) are two trade-offs to optimize. As we increase the number of hidden nodes MSE decreases. However, the increase in the number of hidden-neurons might lead to overfitting the data, thus decreasing the generalization performance of the model. The goal is to find an optimal solution for both model complexity and MSE that maximizes the model performance. We exploited PMOL for estimating the optimal number of parameters for our autoencoder. It uses vector of objective functions and therefore number of optimal solutions are more than one. Pareto front of optimal solution is a set of non-dominated solutions, being chosen as optimal, if no objective can be improved without sacrificing at least one other objective. On the other hand a solution $X$ is referred to as dominated by another solution $Y$ if, and only if, $Y$ is equally good or better than $X$ with respect to all objectives.
Pareto based multi object optimization can be formulated as eq. \ref{par4} with $Q$ objective functions;

\begin{equation}
\label{par4}
f(p) = [f_i(p), i = 1,.....,Q]
\end{equation}
subjected to the 𝐽 equality constraints

\begin{equation}
g_j(p) = 0    \qquad j = 1, 2,..., J
\end{equation}

And the K inequality constraints
\begin{equation}
h_k(p) \le 0 \qquad k = 1,2,...,K
\end{equation}

The aim is to find vector $p^*$ which minimizes $f(p)$, in our case since we have two objective function thus pareto based bi-objective learning problem can be formulated to minimize the two objectives, that is data fitting term and model complexity term given by eq. \ref{par5}

\begin{equation}
\label{par5}
f_1 = -\mathcal{L}(E|\Theta), \qquad f_2 = \gamma k \log(L)
\end{equation}

where $f_1 = -\mathcal{L}(E|\Theta)$ is data fitting objective function and $f_2 = \gamma k \log(L)$ is model complexity objective function. $f_1$  is log-likelihood function that is found with a maximum likelihood estimation algorithm. $E = (\varepsilon_1,\varepsilon_2,...\varepsilon_L)$ is a set of multi-dimensional reconstruction error. Assuming the error is multivariate normal distribution with Mean vector $M=0$ and covariance $\Sigma$, then the distribution function is given by:

\begin{equation}
p(\varepsilon) = \mathcal{N}(M,\Sigma) = \frac{1}{2\pi\sqrt{|\Sigma n_f^2|}} \exp(-\frac{1}{2} \varepsilon^T \Sigma^{-1} \varepsilon)
\end{equation}

where the negative log-likelihood function will be represented by
\begin{equation}
- \mathcal{L}(E|\Theta) = \log \prod_{i=1}^{L} p(\varepsilon_i)
\end{equation}

$\gamma$ is a constant determined by a pareto optimizer, $L$ is input training sample size, $k$ is the number of parameters of the model to be estimated (weight and bias).

\tabcolsep 0.08 in
\begin{table}[t]
	\def\arraystretch{1.5}
	\begin{center}
		\begin{tabular}{|L{3cm}|L{2.6cm}|L{1.7cm}|}
			\hline
			{Methods}       & \textbf{Orange juice} & \textbf{Wine} \\
			\hline\hline
			Douak et al.\cite{douak2011two}     & 01574 & 0.0070   \\ \hline 
			
			Benoudjit et al. \cite{benoudjit2009multiple} & 0.2435 & 0.0052   \\ \hline 
			
			Alhichri et al. \cite{alhichri2013novel} & 0.32076 & 0.0034   \\ \hline 
			
			Ours &  0.1711 & 0.0042   \\ \hline
		\end{tabular}
		\caption{Quantitative results of our method against 3 state-of-the-art methods. Numeric values shows Normalized Mean Square Error (NMSE), the lower is NMSE, the better is performance.}
		\label{table:QA}
	\end{center}
\end{table}


\section{Gaussian Process Regression}
\label{GPR}
Gaussian process regressor (GPR) is a powerful non-parametric technique with precise uncertainty model used in regression problems. Instead of finding correct parameters to fit for basis function, GPR focus on deducing the correlation in all the measured data. A Gaussian process is a Gaussian random function that is characterized by mean and covariance functions, $m(x)$ and $k(x,x')$:
\begin{equation}
f(x) 	\sim GP(m(x), k(x,x'))
\end{equation}
A detailed description GPR is beyond the scope of this paper. Interested readers may refer to \cite{rasmussen2004gaussian}. We used GPR in this work because our data is non-linear and GPR fits very well for our problem. 


\section{Experiments}
\label{EX}
In this work we have used two spectrophotometric datasets, coming from the food industry. The first dataset deals with determining sugar content in the orange juice sample by near infrared reflectance spectrometry \cite{dataset}. The training and test samples in the orange juice data are 149 and 67 respectively; with 700 spectral variables that represents the absorbance ($\log 1/R$) between 1100nm and 2500nm. The value of $R$ represents light reflected by the sample. The concentration of sugar ranges from 0$\%$ to 95.2$\%$ by weight in the sample. The second dataset deals with the determination of alcohol content by mid-infrared spectroscopy in wine samples \cite{dataset}. The training and test data sets contain 91 and 30 spectra, respectively, with 256 spectral variables that are the absorbance ($\log 1/T$) at 256 wave numbers between 4000 and 400 $cm^{-1}$ (where $T$ is the light transmittance through the sample thickness). Alcohol content varies from 7.48$\%$ to 15.5$\%$ by volume. No preprocessing has been performed on the orange juice and wine datasets.

For the quantitative results, the accuracy of the approach is represented in normalized mean square error (NMSE) metric, which is define as:
\begin{equation}
NMSE = \frac{1}{V_{train+test}} \sum_{i=1}^{M} (y_{itest} - y'_{itest})^2
\end{equation}

where $M$ represents the number of testing samples, $y_{itest}$ and $y'_{itest}$ are the real and estimated outputs for the $i^{th}$ test sample $x_{itest}$ and $V_{train+test}$ is the combined variance of the training and test output samples $y_{itrain}$ and $y_{itest}$. The results are listed in Table \ref{V_{train+test}}. It can be seen that our approach gives good results on both the datasets compare to other state-of-the-art methods.

\section{Conclusion}
\label{CN}
We propose a novel deep learning based chemometric data analysis technique for estimating sugar and alcohol concentration in orange juice and wine samples. L2 regularized sparse autoencoder is trained end-to-end for reducing the size of the feature vector to handle the classic problem of curse of dimensionality in chemometric data analysis. The optimal set of parameters of the autoencoder are selected through pareto optimization.  
And Gaussian process regressor is applied on the reduced size feature vector for estimating the concentration of sugar and alcohol in the corresponding samples. The method is compared against 3 state-of-the-art methods. Quantitative results are shown on Normalized Mean Square Error (NMSE) and our approach shows better results on both datasets. 
In future, we are planning to test our approach on more datasets. Moreover, we are also planning to use stacked-autoencoder for feature reduction. Furthermore, we would apply different regressor like linear, support vector and kernel support vector regressor for estimating the corresponding quantities.   



{\small
	\bibliographystyle{IEEEbib}
	\bibliography{Bilal}
}

\end{document}